# A Multi-Swarm Cellular PSO based on Clonal Selection Algorithm in Dynamic Environments

Somayeh Nabizadeh
Computer & IT Engineering Department
Islamic Azad University, Qazvin Branch
Qazvin, Iran
s_nabizadeh@aut.ac.ir

Alireza Rezvanian
Computer & IT Engineering Department
Amirkabir University of Technology
Tehran, Iran
a.rezvanian@aut.ac.ir

Mohammd Reza Meybodi
Computer & IT Engineering Department
Amirkabir University of Technology
Tehran, Iran
meybodi@aut.ac.ir

*Abstract*— **Many real-world problems are dynamic optimization problems. In this case, the optima in the environment change dynamically. Therefore, traditional optimization algorithms disable to track and find optima. In this paper, a new multi-swarm cellular particle swarm optimization based on clonal selection algorithm (CPSOC) is proposed for dynamic environments. In the proposed algorithm, the search space is partitioned into cells by a cellular automaton. Clustered particles in each cell, which make a sub-swarm, are evolved by the particle swarm optimization and clonal selection algorithm. Experimental results on Moving Peaks Benchmark demonstrate the superiority of the CPSOC its popular methods.**

## I. INTRODUCTION

In the real-world problems, many of optimization problems are modeled as dynamic optimization problems, which optima change whole the time. The traditional optimization algorithms are failed in the dynamic environments, because the traditional algorithms cannot track the changes in the environments [1]. Several approaches such as maintenance diversity methods, increase diversity methods, memory-based methods and multi-swarm methods are developed for solving dynamic optimization problems [2].

One of the successful algorithms in dynamic environments is particle swarm optimization (PSO) algorithm [3] with several developments. In [4], the parents maintain diversity and identify the promising regions while the offspring search local areas to find local extrema by multi-swarm algorithm. Cellular PSO is proposed by Hashemi et al [5-6] based on PSO and cellular automata. In cellular PSO, the search space is partitioned into cells. Individual in each cells make a sub-population, are evolved by PSO.

Moreover, Evolutionary algorithms such as genetic algorithm [7], differential evolution [8], artificial immune system [9-10] and ant colony optimization were presented for solving dynamic optimization problems.

In this paper, due to diversity maintenance and speeding up, the hybrid algorithm based on multi-swarm cellular PSO and clonal selection algorithm is proposed. In the proposed algorithm, a simple clustering is used to form sub-population in each cell and clonal selection algorithm is applied for each swarm in each cell to improve solutions.

The rest of this paper is organized as follows: section II briefly introduced the cellular PSO. The proposed algorithm is described in section III. Experimental results of proposed algorithm on Moving Peaks Benchmark as a popular dynamic environment with comparison with alternative algorithms are reported in section IV. Finally, section 5 concludes the paper.

## II. CELLULAR PSO

The original PSO introduced in 1995, is based on swarm behavior by Kennedy and Eberhart. In PSO, each solution is considered as a particle which represents a single bird in a swarm. Initially, the particles are created and positioned randomly in the search space. Then, each particle is updated iteratively according to the best personal and global fitness observed value to reach optimal fitness [11].

In Cellular PSO, the search space is partitioned by a cellular automaton (CA). Particles in each cell in the CA searches and control its corresponding region according to some transmission state rules of CA. Each particle is assigned to a cell based on its position in the space and the search procedure is performed separately for each cell and its neighbors by using the PSO. This search method provides enough diversity and provides the ability of following multiple optimum solutions. In addition, neighboring cells communicate information about their best known solutions which results in a more appropriate cooperation between neighboring cells sharing their experiences. This in turn increases the efficiency of the algorithm [6].

At each iteration in the algorithm, the velocity and position

of the particles are updated according to the following equations:

$$v_i(t+1) = wv_i(t) + c_1 r_1 (pBest_i - p_i(t)) \qquad (1)$$
$$+ c_2 r_2 (lBestMem_k - p_i(t))$$
$$p_i(t+1) = p_i(t) + v_i(t) \quad i = 1,...,m \qquad (2)$$

Where $v_i$ is the velocity of the $i^{th}$ particle and $p_i$ is its position, while $r_1$ and $r_2$ are uniformly distributed random variables in (0,1). $c_1$ and $c_2$ are the learning parameters which are often considered to be equal. $w$ is the inertia weight which may be constant or variable. $pBest_i$ is the best known solution for the $i^{th}$ particle and $lBestMem_k$ is the best known solution of $k^{th}$ cell neighbor to which particle $i$ belongs.

Kamosi et al. proposed a multi-swarm algorithm for dynamic environments, which address the diversity loss problem by introducing two types of swarm: a parent swarm, which explores the search space to find promising area containing local optima and several non-overlapping child swarms, each of which is responsible for exploiting a promising area found by the parent swarm [12].

The main drawback of cellular PSO is that the number of cells exponentially increases as the dimension of the problem and/or the number of partitions increase. Moreover, it is not possible to change the number of cells during runtime. To overcome the problem of fixed number of cells, clustering is used to dynamically create swarms in each cell. Due to clustering technique, it is not necessary to increase the number of cells in order to obtain a more precise search and therefore, the exponential increase in the number of cells is prevented. Moreover, clonal selection algorithm is applied for solution improvement.

### III. CLONAL SELECTION ALGORITHM

Artificial immune system algorithms are bio-inspired algorithms which have been inspired by natural immune system. Natural immune system is divided into innate immune and adaptive immune, the algorithms are proposed by researchers, which are modeled based on the latter. Moreover, based on several theories for natural immune system, the inspired models are developed into five groups: (1) negative selection algorithms, (2) Artificial immune network, (3) clonal selection algorithms, (4) Danger Theory inspired algorithms, (5) Dendritic cell algorithms [13-14].

Clonal selection algorithm (CSA) is one of the artificial immune system algorithms, which is inspired from natural immune system mechanisms. Clonal selection was proposed by de Castro for learning and optimization [15]. The main features of CSA were considered such as affinity proportional reproduction and hypermutation.

The general procedure of CSA is defined as follows:

| Clonal Selection Algorithm (CSA) |
|---|
| 1. Initialize population of antibodies randomly. |
| 2. Repeat steps 3-6 until stopping conditions is met |
| 3. Evaluate the antibodies and calculate their affinity |
| 4. Select a number of the highest affinity antibodies of N and cloning these antibodies with respect their affinities |
| 5. Mutate all the colonies with a rate proportional to their affinities |
| 6. Add the mutated antibodies to the population and reselect a number of the maturated antibodies as memory |

It is note that, several versions of CSA are developed by researchers in the literature, although it is used standard CSA in the proposed algorithm.

### IV. PROPOSED ALGORITHM

In cellular PSO, a CA is used for solution space partitioning. CA technique is known as a mathematical model of systems with several simple components which have local communications. Using CA and the local rules, an ordered structure may be obtained from a completely random state. Two well-known neighborhood structures of Von Neumann and Moore in CA are shown in figure 1.

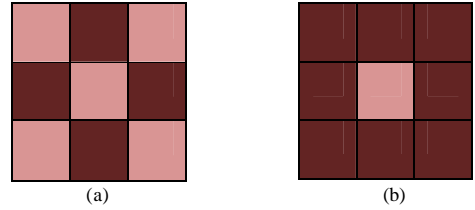

Figure 1. 2-D Neighborhood structure in CA; (a) Moore; (b) Von Neumann

In our proposed method, generally, after partitioning the space into cells, a simple clustering is applied on particles to make swarms in each cell. In during the cellular PSO, clonal selection is used to improve the solutions. In the proposed algorithm, the position and the velocity of the particles are updated according to equation (3) and equation (2).

$$v_i(t+1) = wv_i(t) + c_1 r_1 (pBest_i - x_i(t)) \qquad (3)$$
$$+ c_2 r_2 (sBestMem_k - x_i(t))$$

Where $sBestMem_k$, is the best position in group $k$, is used as dynamic memory of the group to which particle $i$ belongs. In addition, best particles in each group are updated by using the following equation.

$$v_i^k(t+1) = c_1 r_1 (pBest_i - x_i(t)) + c_2 r_2 (cBestMem_k - x_i(t)) \qquad (4)$$
$$+ c_3 r_3 (x_i(t) - sBestMem_l^k) + wv_i^k(t)$$

Where $sBestMem_l^k$, the best position in dynamic memory of the cell to which the group belongs and $sBestMem_l^k$ is the best position in other groups in the cell. As an example, if the best particle in group $j$ of cell $k$ is updated, $sBestMem_l^k$ is the best position among all the groups in cell $k$ except group $j$. The forth part of the Equation prevents the groups from converging to a local extrema.

In the proposed algorithm, after each change clonal selection which increases the efficiency of the algorithm is performed for each swarm. The CSA is applied to the CBest of each cell. The overall process includes definition of a magnitude and a direction of movement for each dimension determining the magnitude and direction of the search in that dimension. Moving in each dimension according to the specified magnitude and direction, the fitness is calculated for the obtained position and if improved, the current position is substituted by the obtained one and if not, the movement direction is reversed in that dimension and a new direction is followed. If the fitness is improved performing the latter action, the update is performed and if not, the magnitude of movement is decreased and the process begins for the next dimension. The whole procedure is performed for all dimensions until in a movement further improvement is impossible in all dimensions and the minimum movement magnitude is reached in all dimensions.

The flow chart of proposed algorithm is illustrated as figure 2.

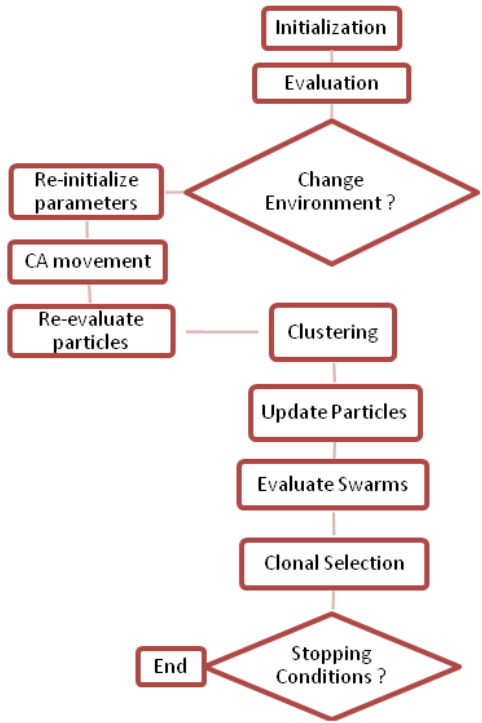

Figure 2. Flow chart of proposed method

In the algorithm, when particles in a swarm converge to a position, the swarm becomes inactive and its particles are used as free particles in finding better solutions in other swarms of the cell or in the neighbor cells. Figure 3 depict the running of the algorithm of particles in a 2-D search space.

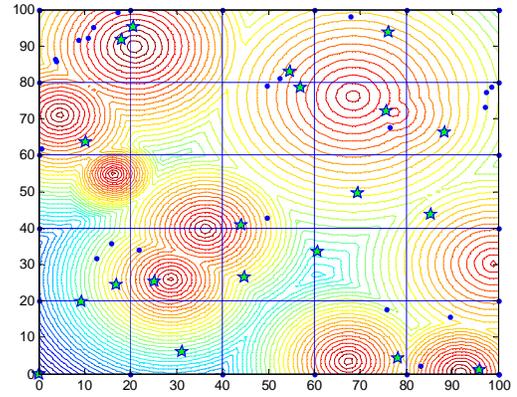

(a)

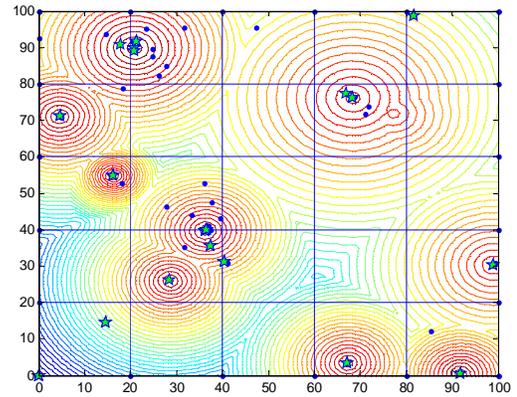

(b)

Figure 3. (a) Particles in the partitioned search space (b) new position of the particles in the search space after some iterations.

## V. EXPERIMENTAL RESULTS

In order to evaluate the proposed algorithm in dynamic environments, several experiments are performed on Moving Peaks Benchmark (MPB). In the MPB, there are some peaks in a multi-dimensional space, where the height, width, and position of each peak alter when the environment changes. Unless stated otherwise, the parameters of the MPB are set to the values presented in table 1 [12,16-17].

TABLE I. DEFAULT SETTINGS OF MPB

| Parameter | Value |
|---|---|
| number of peaks $m$ | 10 |
| Frequency of change $f$ | every 5000 evaluations |
| height severity | 7.0 |
| width severity | 1.0 |
| peak shape | Cone |
| shift length $s$ | 1.0 |
| number of dimensions $D$ | 5 |
| cone height range $H$ | [30.0, 70.0] |
| cone width range $W$ | [1, 12] |
| cone standard height $I$ | 50.0 |
| Search space range $A$ | [0, 100] |

In order to compare the proposed method with other algorithms, the offline error (OE), provided by the equation below, is used.

$$OE = \frac{1}{T}\sum_{t=1}^{T}\left(f\left(p_{best}(t)\right)\right) \quad (5)$$

Where $f$ is the fitness function, $T$ is the maximum number of iterations and $pBest(t)$ is the best known global solution found by the algorithm in iteration $t$.

For the proposed method the inertia weight is considered as a random variable between 0.4 and 0.9. The acceleration coefficient is set to 1.496180, the number of particles is 40; the type of neighborhood structure is Moore and the size of partition is 5.

In the experiments, proposed algorithm so called Cellular PSO based on Clonal Selection as CPSOC is compared with Hibernating Multi Swarm Optimization as HmSO [17], Learning Automata based Immune Algorithm as LAIA [9], Cellular Differential Evolution as CDE [8], Cellular Particle Swarm Optimization as CPSO [6], and Adaptive Particle Swarm Optimization as APSO [16] by offline error. For each experiment, the average offline error and standard deviation of 30 times independent running is reported. The results of several dynamics are listed in the table II, to table V.

TABLE II. OFFLINE ERROR ±STANDARD ERROR FOR F =500

| M | HmSO | LAIA | CDE | CPSO | APSO | CPSOC |
|---|---|---|---|---|---|---|
| 1 | 8.53±0.49 | 7.34±0.32 | 8.20±0.19 | 7.81±0.51 | 4.81±0.14 | 8.29±0.55 |
| 5 | 7.40±0.31 | 7.05±0.39 | 6.06±0.05 | 6.59±0.31 | 4.95±0.11 | 6.29±0.21 |
| 10 | 7.56±0.27 | 6.91±0.32 | 5.93±0.04 | 7.35±0.22 | 5.16±0.11 | 5.45±0.17 |
| 20 | 7.81±0.20 | 6.95±0.38 | 5.60±0.03 | 7.79±0.27 | 5.81±0.08 | 5.47±0.19 |
| 30 | 8.33±0.18 | 6.92±0.33 | 5.56±0.03 | 7.88±0.23 | 6.03±0.07 | 5.59±0.12 |
| 40 | 8.45±0.18 | 6.84±0.31 | 5.47±0.02 | 7.83±0.21 | 6.10±0.08 | 5.63±0.16 |
| 50 | 8.83±0.17 | 6.43±0.29 | 5.47±0.02 | 8.12±0.22 | 5.95±0.06 | 5.74±0.11 |
| 100 | 8.85±0.16 | 6.58±0.26 | 5.29±0.02 | 7.90±0.24 | 6.08±0.06 | 5.45±0.07 |
| 200 | 8.85±0.16 | 6.41±0.27 | 5.07±0.02 | 7.82±0.20 | 6.20±0.04 | 5.79±0.10 |

TABLE III. OFFLINE ERROR ±STANDARD ERROR FOR F =1000

| M | HmSO | LAIA | CDE | CPSO | APSO | CPSOC |
|---|---|---|---|---|---|---|
| 1 | 4.46±0.26 | 4.96±0.32 | 4.98±0.35 | 5.86±0.42 | 2.72±0.04 | 4.74±0.32 |
| 5 | 4.27±0.08 | 4.01±0.31 | 3.96±0.04 | 5.26±0.26 | 2.99±0.09 | 3.95±0.21 |
| 10 | 4.61±0.07 | 3.94±0.29 | 3.98±0.03 | 5.75±0.23 | 3.87±0.08 | 3.20±0.20 |
| 20 | 4.66±0.12 | 3.72±0.29 | 4.53±0.02 | 5.74±0.19 | 4.13±0.06 | 3.52±0.17 |
| 30 | 4.83±0.09 | 4.03±0.31 | 4.77±0.02 | 5.84±0.16 | 4.12±0.04 | 3.96±0.12 |
| 40 | 4.82±0.09 | 3.97±0.32 | 4.87±0.02 | 5.84±0.17 | 4.15±0.04 | 4.21±0.17 |
| 50 | 4.96±0.03 | 4.22±0.31 | 4.87±0.02 | 5.84±0.14 | 4.11±0.03 | 3.98±0.11 |
| 100 | 5.14±0.08 | 4.19±0.32 | 4.85±0.02 | 5.73±0.11 | 4.26±0.04 | 4.13±0.12 |
| 200 | 5.25±0.08 | 4.38±0.31 | 4.46±0.01 | 5.48±0.11 | 4.21±0.02 | 4.15±0.01 |

TABLE IV. OFFLINE ERROR ±STANDARD ERROR FOR F =2500

| M | HmSO | LAIA | CDE | CPSO | APSO | CPSOC |
|---|---|---|---|---|---|---|
| 1 | 1.75±0.10 | 2.48±0.15 | 2.38±0.78 | 3.78±0.25 | 1.06±0.03 | 2.31±0.21 |
| 5 | 1.92±0.11 | 2.51±0.19 | 2.12±0.02 | 2.91±0.14 | 1.55±0.05 | 2.01±0.13 |
| 10 | 2.39±0.16 | 2.82±0.27 | 2.42±0.02 | 3.18±0.16 | 2.17±0.07 | 1.56±0.15 |
| 20 | 2.46±0.09 | 3.16±0.36 | 3.05±0.04 | 3.65±0.13 | 2.51±0.05 | 2.41±0.13 |
| 30 | 2.57±0.05 | 3.14±0.33 | 3.29±0.03 | 3.90±0.11 | 2.61±0.02 | 2.78±0.10 |
| 40 | 2.56±0.06 | 3.02±0.31 | 3.43±0.03 | 4.20±0.13 | 2.59±0.03 | 2.90±0.12 |
| 50 | 2.65±0.05 | 3.05±0.31 | 3.44±0.02 | 4.08±0.11 | 2.66±0.02 | 3.18±0.09 |
| 100 | 2.72±0.04 | 3.14±0.35 | 3.36±0.01 | 4.23±0.09 | 2.62±0.02 | 3.22±0.07 |
| 200 | 2.81±0.04 | 3.08±0.32 | 3.13±0.01 | 4.09±0.10 | 2.64±0.01 | 3.09±0.12 |

TABLE V. OFFLINE ERROR ±STANDARD ERROR FOR F =5000

| M | HmSO | LAIA | CDE | CPSO | APSO | CPSOC |
|---|---|---|---|---|---|---|
| 1 | 0.87±0.05 | 1.94±0.19 | 1.53±0.07 | 2.36±0.14 | 0.53±0.01 | 1.02±0.14 |
| 5 | 1.18±0.04 | 2.09±0.18 | 1.50±0.04 | 1.94±0.16 | 1.05±0.06 | 0.99±0.15 |
| 10 | 1.42±0.04 | 2.14±0.15 | 1.64±0.03 | 2.09±0.13 | 1.31±0.03 | 1.75±0.10 |
| 20 | 1.50±0.06 | 2.97±0.21 | 2.64±0.05 | 2.94±0.13 | 1.69±0.05 | 1.93±0.11 |
| 30 | 1.65±0.04 | 2.98±0.23 | 2.62±0.05 | 3.04±0.09 | 1.78±0.02 | 2.28±0.10 |
| 40 | 1.65±0.05 | 3.07±0.29 | 2.76±0.05 | 3.16±0.11 | 1.86±0.02 | 2.62±0.09 |
| 50 | 1.66±0.02 | 2.93±0.27 | 2.75±0.05 | 3.19±0.10 | 1.95±0.02 | 2.74±0.10 |
| 100 | 1.68±0.03 | 3.06±0.24 | 2.73±0.03 | 3.24±0.09 | 1.95±0.01 | 2.84±0.12 |
| 200 | 1.71±0.02 | 2.95±0.23 | 2.61±0.02 | 3.15±0.08 | 1.90±0.01 | 2.69±0.08 |

According to the results of the table II to table V, the proposed algorithm is better than alternative algorithms relatively.

## VI. CONCLUSION

In this paper, an improving cellular PSO algorithm proposed by clonal selection algorithm in dynamic environment. The proposed algorithm, speed up the clellular PSO algorithm by clustering and improve the quality of solutions by clonal selection algorithm. Moreover in the proposed algorithm, it is applied precise search without increasing the number of partitions. The experimental results on the moving peaks benchmark as a popular dynamic environment with comparison with state of the art algorithms shows the relative improvements for our suggestions.